\documentclass[conference]{IEEEtran}
\IEEEoverridecommandlockouts
\usepackage{cite}
\usepackage{amsmath,amssymb,amsfonts}
\usepackage{algorithmic}

\usepackage{algorithm}
\usepackage{graphicx}
\usepackage{textcomp}
\usepackage{xcolor}
\usepackage{mathtools}
\usepackage{caption}
\usepackage{subcaption}

\def\BibTeX{{\rm B\kern-.05em{\sc i\kern-.025em b}\kern-.08em
    T\kern-.1667em\lower.7ex\hbox{E}\kern-.125emX}}
\begin{document}

\title{Federated Semi-Supervised Domain Adaptation via Knowledge Transfer}

\author{\IEEEauthorblockN{Madhureeta Das\textsuperscript{1}, Xianhao Chen\textsuperscript{2}, Xiaoyong Yuan\textsuperscript{3}, Lan Zhang\textsuperscript{1}}
\IEEEauthorblockA{\textsuperscript{1}{Department of Electrical and Computer Engineering}, Michigan Technological University, Houghton, MI, USA\\
\textsuperscript{2}Department of Electrical and Electronic Engineering, University of Hong Kong, Hong Kong, China\\
\textsuperscript{3}{College of Computing}, Michigan Technological University, Houghton, MI, USA \\
\{mdas1, xyyuan, lanzhang\}@mtu.edu, xchen@eee.hku.hk}
}

\maketitle

\begin{abstract}
Given the rapidly changing machine learning environments and expensive data labeling, semi-supervised domain adaptation (SSDA) is imperative when the labeled data from the source domain is statistically different from the partially labeled data from the target domain. Most prior SSDA research is centrally performed, requiring access to both source and target data. However, data in many fields nowadays is generated by distributed end devices.  Due to privacy concerns, the data might be locally stored and cannot be shared, resulting in the ineffectiveness of existing SSDA research. This paper proposes an innovative approach to achieve SSDA over multiple distributed and confidential datasets, named by Federated Semi-Supervised Domain Adaptation (FSSDA). FSSDA integrates SSDA with federated learning based on strategically designed knowledge distillation techniques, whose efficiency is improved by performing source and target training in parallel. Moreover, FSSDA controls the amount of knowledge transferred across domains by properly selecting a key parameter, \textit{i.e.}, the imitation parameter. Further, the proposed FSSDA can be effectively generalized to multi-source domain adaptation scenarios. Extensive experiments are conducted to demonstrate the effectiveness and efficiency of FSSDA design.

\end{abstract}

% \begin{IEEEkeywords}
% component, formatting, style, styling, insert
% \end{IEEEkeywords}

\section{Introduction}\label{Introduction}
Domain adaptation addresses a common situation that arises when applying machine learning to diverse data~\cite{wang2018deep}. Taking object detection as an example, a model trained for autonomous driving using data from sunny weather may perform poorly on foggy or snowy days. 
Typically, there is ample labeled data from the source domain to train the original model (e.g., sunny day object detection) but little labeled data from the target domain for domain adaptation (e.g., snowy day object detection). Given the fast-changing machine learning environments and expensive data labeling, it is critical to develop domain adaptation approaches to handle the domain shift when there is abundant unlabeled data and limited labeled data from the target domain, \textit{i.e.}, semi-supervised domain adaptation (SSDA).

Prior SSDA research is mainly conducted in a centralized manner, requiring access to both source and target domain data~\cite{daume2010frustratingly,donahue2013semi,yao2015semi}. 
However, data in many fields nowadays is generated by distributed end devices, such as mobile and IoT devices. Given the widespread impact of recent data breaches~\cite{DataBreach21}, end users may become reluctant to share their local data due to privacy concerns. Although federated learning (FL)~\cite{yang2019federated} offers a promising way to enable knowledge sharing across end devices without migrating the private end data to a central server, it is non-trivial to marry existing SSDA approaches with FL paradigm. First, data from both the source and target domains is stored at end devices and cannot be shared in federated settings, resulting in the ineffectiveness of the existing centralized SSDA. Second, efficiency has been a well-recognized concern for FL. With distributed data from both source and target domains, more iterations need to be involved in obtaining a well-trained target model. Last but not least, the entangled knowledge across domains may lead to negative transfer~\cite{pan2009survey}, which becomes more challenging in federated settings with unavailable data from source and target domains across devices. 

Enlighten by a popular model fusion approach, knowledge distillation (KD), that allows knowledge transfer across different models~\cite{hinton2015distilling}, we enable knowledge transfer between models from different domains without accessing the original domain data. 
Specifically, the target model can be learned with the help of the soft labels that are predictions of target samples by using the source model. Considering the distributed data from both source and target domains in federated settings, instead of waiting for a well-trained source model, we propose a parallel training paradigm to generate soft labels along with the source model to improve SSDA efficiency. However, due to domain discrepancy, the soft labels generated from the source model can be different from the ground truth target labels. Moreover, the soft labels derived at the initial federated training stage may perform poorly on SSDA. To address the above issues, we intend to align the source and target domains by adaptively leveraging both soft labels and the ground truth labels. One major challenge here is the limited ground truth target labels in SSDA. To effectively leverage the few ground truth labels, we balance the knowledge transferred from the soft and ground truth labels by properly selecting a key parameter, \textit{i.e.}, the imitation parameter. Inspired by recent multi-task learning research~\cite{milojkovic2019multi}, we control the amount of knowledge transferred from the source domain by adaptively selecting the imitation parameter based on the stochastic multi-subgradient descent algorithm (SMSGDA). The adaptively derived imitation parameters can be effectively used to handle multi-source SSDA problems under federated settings.

By integrating the above ideas, we propose an innovative SSDA approach for federated settings, named Federated Semi-Supervised Domain Adaptation (FSSDA). To the best of our knowledge, the research we present here is the first SSDA approach over distributed and confidential datasets. Our main contributions are summarized as follows: 
\begin{itemize}
    \item To achieve SSDA over multiple distributed and confidential datasets, we propose FSSDA to integrate SSDA and FL, which enables knowledge transfer between source domain(s) and target domain by leveraging domain models rather than original domain data based on strategically designed knowledge distillation techniques.
    \item Considering distributed data from both source and target domains in federated settings, we develop a parallel training paradigm to facilitate domain knowledge generation and domain adaptation concurrently, improving the efficiency of FSSDA.
    \item Due to different domain gaps in various SSDA problems, we control the amount of knowledge transferred from different domains to avoid negative transfer, where the imitation parameter, a key parameter of FSSDA, is properly selected based on SMSGDA algorithm. 
    \item Extensive experiments are conducted on the office dataset under both iid and non-iid federated environments. Experimental results validate the effectiveness and efficiency of the proposed FSSDA approach. 
\end{itemize}

This paper is organized as below. Section~\ref{sec:rel} reviews the related work. Section~\ref{sec:FSSDA} describes the FSSDA approach and key modules of the FSSDA design. Section~\ref{sec:exp} describes the experimental framework and evaluation results. Finally, in section~\ref{sec:con} we conclude the overall work.
 
\vspace{0.5em}
\section{Related Work}\label{sec:rel}
\subsection{Semi-Supervised Domain Adaptation (SSDA)}

SSDA intends to address the domain shift when the labeled data from the source domain is statistically different from the partially labeled data from the target domain~\cite{wang2018deep}. Classical SSDA research explots the knowledge from the source domain by mitigating the domain discrepancy~\cite{daume2010frustratingly,donahue2013semi,yao2015semi}. Daum\'e \textit{et al.}~\cite{daume2010frustratingly} proposed to compensate the domain discrepancy by augmenting the feature space of source and target domain data. Donahue \textit{et al.}~\cite{donahue2013semi} solved the domain discrepancy problem by optimizing the auxiliary constraints on the labeled data. Yao \textit{et al.}~\cite{yao2015semi} proposed a framework named semi-supervised domain adaptive subspace learning (SDASL) to learn a subspace that can reduce the data distribution mismatch. 
Saito \textit{et al.}~\cite{saito2019semi} minimize the distance between the unlabeled target samples and the class prototypes through minimax training on entropy. Some recent research proposed adversarial-based methods, such as DANN~\cite{ganin2016domain}, to adversarially learn discriminative and domain-invariant representations. 
However, all above SSDA research requires access to both source and target domain data. Although in one recent work GDSDA~\cite{ao2017fast}, the source domain data requirement is relaxed, GDSDA is designed to learn a shallow target model, \textit{i.e.}, the SVM. Meanwhile, target domain samples are still required to estimate target domain labeling for SSDA supervision, leading to the ineffectiveness for deep learning-based SSDA over distributed and confidential datasets from both source and target domains.

\subsection{Label-Limited Federated Learning (FL)} 
FL has been a popular learning paradigm for knowledge transfer across distributed and confidential datasets.

Most of the existing FL research focuses on supervised learning tasks with ground-truth labeled samples at end devices. However, the end data is often unlabeled in practice, since annotating requires both time and domain knowledge~\cite{zhu2009introduction,zhang2021fedzkt}. 
Some recent research has focused on the label-limited FL problems, mainly about semi-supervised FL and unsupervised domain adaptation (UDA) under federated settings. 
To handle semi-supervised FL, Albaseer \textit{et al.}~\cite{albaseer2020exploiting} proposed FedSem by developing distributed processing schemes based on pseudo-labeling techniques. Similarly, Jeong \textit{et al.}~\cite{jeong2020federated} introduced the inter-client consistency loss to transfer labeling knowledge from the labeled samples to nearby unlabeled ones with high confidence. 
Another line of label-limited FL research on UDA problems become more challenging due to the data requirements in prior UDA research~\cite{wang2018deep}. Peng \textit{et al.}~\cite{peng2019federated} proposed FADA to transfer source knowledge across multiple distributed nodes to a new target node by using adversarial-based approaches. Peterson \textit{et al.}~\cite{peterson2019private} leveraged a prior domain expert to guide per-user domain adaptation. Zhuang \textit{et al.}~\cite{zhuang2021towards} predict pseudo labels by developing a new clustering algorithms for individual face recognition. On the one hand, the above research targets either a single source dataset or a single target dataset 
for personalization or knowledge aggregation, while our design is under multiple distributed source and target datasets for a more general domain adaptation setting. 
On the other hard, UDA problems assume unknown target ground-truth labels, which cannot effectively extract target domain knowledge from the few labeled target domain samples in SSDA.

\subsection{Knowledge Distillation (KD)}
KD was initially proposed to compress a large neural model (teacher model) down to a smaller model (student model)~\cite{hinton2015distilling,chen2020online}. Typically, KD compresses the well-trained teacher model into an empty student model by steering the student's prediction towards teacher's prediction~\cite{polino2018model}.  
Urban \textit{et al.}~\cite{urban2016deep} used a small network to simulate the output of large depths using layer-by-layer distillation. Similarly, \cite{luo2016face} uses $\ell_2$ loss to train a compressed student model from a teacher model for face recognition. Previous works ~\cite{yim2017gift,bagherinezhad2018label,furlanello2018born} also show distilling a teacher model into a student model of the same architecture can improve student over teacher. Furlanello \textit{et al.}~\cite{furlanello2018born} and Bagherinezhad \textit{et al.}~\cite{bagherinezhad2018label} demonstrate that by training the student using softmax outputs of the teacher as ground truth over generations.
Some recent works~\cite{ao2017fast,meng2018adversarial,zhou2021domain} use KD to address domain adaptation problems through a teacher-student training strategy: train multiple teacher models on the source domain and integrate them into the target domain to train the student model. 
However, the above KD-based domain adaptation research requires access either the source or target domain data, which cannot be used to solve SSDA over multiple distributed and confidential datasets from both source and target domains.

\begin{figure}[!tb]
\centering
\includegraphics[width=1\linewidth]{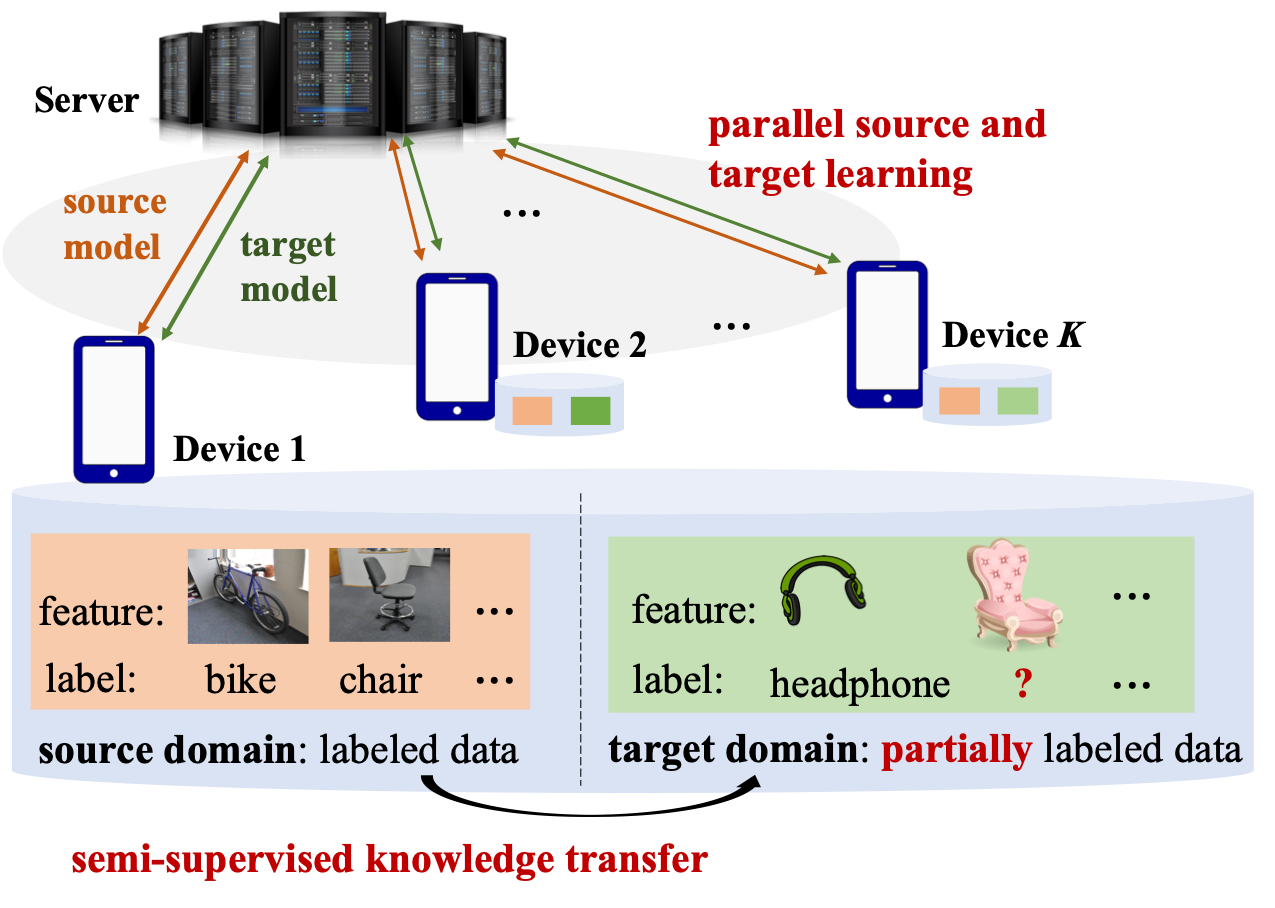}%
\caption{Proposed FSSDA Framework.}
\label{fig:system}
\end{figure}

\section{FSSDA: Federated Semi-Supervised Domain Adaptation}\label{sec:FSSDA}
\subsection{Problem Statement}\label{AA}
This work focuses on a typical SSDA problem over distributed $K$ confidential datasets. Each dataset $\mathcal{D}^k=\{\mathcal{D}_s^k, \mathcal{D}_t^k\}$ includes data from two domains, which is held by an end devices $k$ in a set of $\mathcal{K}$, $|\mathcal{K}|=K$. Specifically, the source domain data at device $k\in\mathcal{K}$ is fully labeled and denoted by $\mathcal{D}_s^k= (\mathcal{X}_s^k,\mathcal{Y}_s^k)$; the target domain data is partially labeled and denoted by $\mathcal{D}_t^k= \{\mathcal{D}^k_{t_l}, \mathcal{D}^k_{t_u} \}$. In particular, the labeled target data $\mathcal{D}^k_{t_l}=(\mathcal{X}_{t_l}^k, \mathcal{Y}_{t_l}^k)$ is much less than the unlabeled target data $\mathcal{D}^k_{t_u}=(\mathcal{X}_{t_u}^k)$. The datasets cannot be shared.  The ultimate \textit{goal} of this work is to obtain a global target model $W_{t}$  that performs well on the distributed target domain data $\mathcal{D}_t = \{\mathcal{D}_t^k\}_{k\in\mathcal{K}}$ without accessing any data from both source and target domains $\mathcal{D} = \{\mathcal{D}^k\}_{k\in\mathcal{K}}$. 

\subsection{FSSDA Design}
To achieve the above goal, we propose an innovative approach named Federated Semi-Supervised Domain Adaptation (FSSDA), as shown in Fig.~\ref{fig:system}. Specifically, we first develop the semi-supervised knowledge transfer module to integrate SSDA with federated learning based on knowledge distillation techniques. Besides, to improve the efficiency of FSSDA, the parallel training module is then proposed to enable concurrent training between source and target domains. Finally, a key parameter of FSSDA, \textit{i.e.}, the imitation parameter, is improved through the imitation parameter selection module to further boost the domain adaptation performance. The overall procedures of FSSDA are illustrated in Algorithm~\ref{alg:fssda}. In the following, we elaborate on the key modules of FSSDA design, respectively.

\vspace{0.5em}
\begin{algorithm}[!tb]
\caption{Federated Semi-Supervised Domain Adaptation (FSSDA) }\label{alg:fssda}
\textbf{INPUT:}  for each device $k\in\mathcal{K}$, source domain data $\mathcal{D}_s^k=(\mathcal{X}_s^k,\mathcal{Y}_s^k)$ and target domain data $\mathcal{D}_t^k=\{(\mathcal{X}^k_{t_l},\mathcal{Y}^k_{t_l}),(\mathcal{X}^k_{t_u})\}$; the number of rounds $R$.
\begin{algorithmic}[1]
\STATE initialize the global source and target model as $W_s(0)$ and $W_t(0)$;
\STATE initialize the local source and target model as $w_s^k(0)$ and $w_t^k(0)$ for each device $k\in\mathcal{K}$;
\FOR{each round $r=1,2,...R$}
\vspace{0.2em}
\STATE // \textit{device source domain update}
\FOR{each device $k\in\mathcal{K}$}
    \STATE $w_s^k(r) \leftarrow$ $DeviceSource(W_s(r-1))$
    
\ENDFOR   
\vspace{0.2em}
\STATE // \textit{server update}
%\State $W_s(r) \leftarrow$ $\sum_{m\in\mathcal{M}} w_s^m(r)$
\STATE $W_s(r) \leftarrow$ $\frac{1}{k} \sum_{i=1}^{k} w_s^i(r)$
\vspace{0.2em}
\STATE // \textit{device target domain update}
\FOR{each device $k\in\mathcal{K}$}
     \STATE $w_t^k(r) \leftarrow$ $DeviceTarget (W_t(r-1),W_s(r))$
     
\ENDFOR
\vspace{0.2em}
\STATE // \textit{server update}

\STATE $W_t(r) \leftarrow$ $\frac{1}{k} \sum_{i=1}^{k} w_t^i(r)$
\ENDFOR
\end{algorithmic}
\end{algorithm}%\vspace{-0.5em}
\subsubsection{Semi-Supervised Knowledge Transfer}
Federated Learning (FL) has been one promising approach to enable knowledge transfer across distributed devices without accessing the private device data~\cite{yang2019federated}. By exchanging the locally trained models, classical FL obtains a global model by element-wise averaging the local models. However, existing FL research has not focused on SSDA problems. Enlighten by the concept of knowledge distillation (KD)~\cite{hinton2015distilling,cho2019efficacy} that distills knowledge from one or more models (teacher) into a new one (student), we develop FSSDA to integrate SSDA and FL based on strategically designed KD techniques to transfer knowledge between devices and domains. 

Typically, KD generates the student model by mimicking the outputs of the teacher model on the same dataset. The dataset here is not necessarily the one on which the teacher model was trained. KD has been adopted to exploit information of unlabeled data~\cite{lopez2015unifying,ao2017fast}. In FSSDA, KD is used to exploit the knowledge of unlabeled target data, where the source model is the teacher and the target model is the student. To enable SSDA, FSSDA assigns each sample in the target domain with a hard label $y_t$ and a soft label $y_t^*$. The hard label for a labeled target sample is its actual label in the one-hot manner. For an unlabeled target sample, we use a ``fake label'' strategy that assigns all $0s$ as the label. Thus, all samples in the target domain have the hard labels. It should be mentioned that although the fake label may introduce some noise, this noise can be controlled by the properly designed imitation parameters by balancing the impact from the hard and soft labels. The similar findings have been found in recent research~\cite{ao2017fast}. The soft label of a target sample is derived by the prediction of the source model, which is a class probability value. By leveraging the source data and the target data with hard and soft labels, the process to train the target model is as follows: 
\begin{itemize}
    \item  Train the source model $w_s^k$ for device $k\in\mathcal{K}$ with $\mathcal{D}_s^k$;
    \item Use the learned source model to generate the soft label $y_t^*$ for each sample $x_t\in\mathcal{X}_t$ in the target domain using softmax function $\sigma$. The soft label is defined by %\vspace{-1em}
    \begin{align}  \label{eq:1}
    y_t^*&=\sigma (W_{s} (x_t) / T),%\vspace{-1em}
  \end{align}
  where $W_s$ is the global source model by element-wise averaging local source model $w_s^k$ for all device $k\in\mathcal{K}$~\cite{mcmahan2017communication}, and $T$ is the temperature parameter to control the smoothness of the soft label.
  \item Train the target model $w_t^k$ at device $k$ using the hard and soft labels for each target data by
\begin{flalign}\label{eq:2}\vspace{-0.5em}
  \arg \min \quad \frac{1}{N_t^k}\sum_{i=1}^{N_t^k}[ & \lambda^{k} \ell_t(y_t^i,w_t^k(x_t^i))  &\\ \nonumber 
  &+ (1-\lambda^{k}) \ell_t(y_t^{*i},w_t^k(x_t^i))],\vspace{-1em}
\end{flalign}
  where $N_t^k$ denotes the number of target domain samples at device $k$; $\ell_t$ is the loss function; $w_t$ is the local target model; $\lambda^k$ is the imitation parameter for device $k$ to balance the importance between the hard label $y_t$ and the soft label $y_t^*$. 
\end{itemize}

\begin{algorithm}[!tb]
\caption{FSSDA: \textit{DeviceSource}}\label{alg:client}
\textbf{INPUT:} source domain data $\mathcal{D}_s= (\mathcal{X}_s,\mathcal{Y}_s)$, Source model weight $W_s$, number of local training epoch $E$, local learning rate $\eta$
\begin{algorithmic}[1]
\STATE $w_s=W_s$ 
\FOR{$e \in \{1,2,...,E\}$}
     \STATE $\mathcal{L}_s \gets$
      $\mathcal{L}_{CE}(\mathcal{Y}_s,w_s(\mathcal{X}_s))$
      \STATE $w_s \gets w_s - \eta \frac{\partial \mathcal{L}_s}{\partial w_s}$
\ENDFOR
\end{algorithmic}
\textbf{RETURN: $w_s$}
\end{algorithm}

\begin{algorithm}[!tb]
\caption{FSSDA: \textit{DeviceTarget}}\label{alg:client}
\textbf{INPUT:} target domain data $\mathcal{D}_t$ of size $N_t$, source model weight $W_s$, target model weight $W_t$, number of local training epoch $E$, local learning rate $\eta$
\begin{algorithmic}[1]
\STATE $w_t = W_t$
\FOR{$i \in D_t$}
     \STATE $y^*_i = \sigma (W_s (x_i) / T)$
\ENDFOR
\FOR{$e \in \{1,2,...,E\}$}
     \STATE compute $\lambda$ using equation~\eqref{eq:4}
     \STATE $\mathcal{L}_t \gets$  
      $\frac{1}{N_t}\sum_{i=1}^{N_t}$
      $[ \lambda \mathcal{L}_t(y_t^{*i},w_t(x_t^{i})) + (1 - \lambda) \mathcal{L}_t(y_t^{*i},w_t(x_t^{i}))] $
      \STATE $w_t \gets w_t - \eta \frac{\partial \mathcal{L}_t}{\partial w^t}$
\ENDFOR
\vspace{0.2em}
\end{algorithmic}
\textbf{RETURN: $w_t$}
\end{algorithm}

\vspace{0.5em}
\subsubsection{Parallel Training Paradigm between Source and Target Domains}
Efficiency has been a well-known concern in distributed machine learning. Since both source and target data are distributed across devices, instead of waiting for a well-trained source model, we propose a parallel training paradigm to accelerate FSSDA. As shown in lines 5 and 11 in Algorithm~\ref{alg:fssda}, each device trains the source and target model simultaneously, where the source model is trained by following classical FL process (represented by FedAvg~\cite{mcmahan2017communication}) and the target model is trained based on the process in the above module. Although the source model does not perform well in the initial stage, it still promotes domain alignment and thus accelerates the generation of the target model. Thus the main purpose of parallel computing is to train the source and target models simultaneously, speeding up the overall training process. Our parallel design is empirically evaluated in the experiment section below. It should be mentioned that parallel training does not incur additional communication cost, since target model updates can be appended to source updates.

\vspace{0.5em}
\subsubsection{Adaptive Imitation Parameters}\label{sec:lamb}
FSSDA transfers knowledge from the source domain to the target domain by leveraging both soft labels and hard labels for domain adaptation. 
However, most hard labels are the fake ones in semi-supervised environments, providing limited knowledge from the target domain. Meanwhile, due to the domain gap between source and target, the entangled knowledge learned from the source domain may generate improper soft labels and lead to negative transfer~\cite{pan2009survey}. Such problems can be more serious in federated settings, where target devices do not have access to any source domain data.
To properly balance the importance between hard labels and soft labels, we develop an adaptive approach for selecting the imitation parameter $\lambda$ in equation.~\eqref{eq:2}. 
Specifically, the imitation parameter controls how much knowledge can be transferred from the source domain, whose importance has been shown in prior research~\cite{duan2011visual,duan2012learning,lopez2015unifying}. However, prior research determines the imitation parameter using either a brute-force search or domain knowledge, which cannot flexibly optimize knowledge transfer for difference domain discrepancy under various SSDA problems. Such challenge becomes more significant under heterogeneous federated settings, where end devices have statistically heterogeneous data (non-iid) for both source and target domains.

To effectively select imitation parameters to simultaneously reduce the losses from the soft and hard label, problem (2) can be considered as a multi-task learning problem, where the soft loss and hard loss are the two task objectives. Since in each federated training iteration, each device holds its own target domain data and the global source domain model updated from the server, imitation parameters can be determined independently at the device side, which also address the aforementioned data heterogeneity concern in federated setting. Specifically, we propose to leverage stochastic multi-subgradient descent algorithm (SMSGDA)~\cite{milojkovic2019multi}, a well-known multi-task learning approach, to adaptively select the imitation parameter at each federated iteration for each individual device. In this way, FSSDA can determine the optimal imitation parameter and accelerate the overall training process.  
Specifically, each device computes the gradient of the corresponding losses, and then separates the gradients used by the model optimizer for every single loss $\nabla\ell$. The objective function for imitation parameter selection at each device can be given by
\begin{align}\label{eq:3}\vspace{-1em}
\begin{split}
    \min_{\lambda \in [0,1]} \|\lambda \nabla_w \ell_t(y_t,w_t(x_t)) +  &  (1-\lambda)\nabla_w \ell_t(y_t^{*},w_t(x_t)) \|^2,\vspace{-1em}
\end{split}
\end{align}
where $\ell_t$ is the loss function, $y_t$ is the hard label, and $y_t^{*}$ is the soft label generated by the source model $W_s$ for the target domain dataset. $w_t$ is the local target model. The analytical solution of the above problem can be given by
\begin{align}\label{eq:4}\hspace{-1em}
\begin{split}
    \lambda = \nabla_w & \ell_t( y_t^{*},w_t(x_t))\\
    &{\times \dfrac {(\nabla_w \ell_t(y_t^{*},w_t(x_t)) - \nabla_w \ell_t(y_t,w_t(x_t)))^T } {\| \nabla_w \ell_t(y_t,w_t(x_t)) - \nabla_w \ell_t(y_t^{*},w_t(x_t)) \|^2}},
\end{split}
\end{align}
where  $\lambda$ is clipped between [0,1].
Therefore, 
each device can efficiently derive its local imitation parameter with the above closed-form solution.

\subsection{FSSDA over Multi-Source Domains}
Our prior design mainly targets single-source SSDA problems. This part will introduce the extension of the proposed FSSDA to multi-source scenarios. Specifically, when the distributed source data includes multiple source domains, it is essential to extract the inter-domain knowledge to better align the domain-specific representations. Define the total number of source domains by $S$. Following the aforementioned parallel training paradigm, $S$ source models will be simultaneously trained under federated settings, which respectively provide $S$ soft labels for each single target sample. Thus, the overall learning objective at device $k\in\mathcal{K}$ for $S$ source domains can be extended from (2) to 

\begin{flalign}\label{eq:5}\vspace{-1em}
    \arg & \min  \hspace{-1em} &\\ \nonumber
     & \frac{1}{N_t^k}  \sum_{i=1}^{N_t^k}[  \lambda_1^{k} \ell_t(y_t^i,w_t^k(x_t^i))+\sum_{j=1}^{S} \lambda_{j+1}^{k}  \ell_t(y_t^{*ij},w_t^k(x_t^i))],  \\ \nonumber
     s.t. & \quad \sum\lambda_{i}^{k} = 1,%\vspace{-1em}
\end{flalign}
where $N_t^k$ is the total number of data samples in the target domain at device $k$, $w_t$ is the local target model, $y_t^{*ij}$ is the soft-label generated by the $j$th source model $W_S^j$ for local data $x_i$, and $\lambda^k$ is the imitation parameter for each device $k$.
Likewise, in multi-source scenarios, we need to find proper imitation parameters to control the amount of knowledge transferred from different source domains. As shown in Equation (5), imitation parameters are used to control more than two objective functions, \textit{i.e.}, in total $S+1$ losses, to jointly optimize the target model. Thus, given the new condition for imitation parameters, the constrained optimization problem (5) cannot be solved by the closed-form solution in (4) derived for problem (2). Following the study in~\cite{sener2018multi}, we use Frank-Wolfe to solve the constrained optimization in (5). Moreover, the Frank-Wolfe-based optimizer can scale to high-dimensional problems with low computational overhead~\cite{frank1956algorithm,sener2018multi}.

\vspace{0.5em}
\section{Experiments}\label{sec:exp}
\subsection{Experimental Setup}
We evaluate our models on the office dataset~\cite{saenko2010adapting}, which is widely used in domain adaptation. The office dataset includes 3 subsets: Webcam ($795$ samples) contains images captured by the web camera, Amazon ($2,817$ samples) contains images downloaded from amazon.com, and DSLR ($498$ samples) contains images captured by a digital SLR camera under different photographic settings, sharing 31 classes. 
We consider both iid and non-iid data distributions across the devices in federated settings. We use the distribution-based label imbalance~\cite{li2021federated} to generate non-iid data distributions, where each end device is allocated a proportion of the samples, whose labels follow Dirichlet distribution. Specifically, we sample $p_l$ $\sim$ DirN ($\beta$) and allocate a $p_{l,k}$ proportion of the instances of class l to each device k. Here Dir (·) denotes the Dirichlet distribution, and $\beta$ is a concentration parameter. We can flexibly change the imbalance level by varying the concentration parameter $\beta$. In our setting, we set the $\beta$ value as $0.1$.
We consider practical SSDA settings, where limited labeled samples are given in the target domain. In iid and non-iid settings, only 93 labeled examples (3 per class) are distributed across all the end devices. We use ResNet-101~\cite{he2016deep} for the baseline methods and the proposed method. All the models are pre-trained on the ImageNet~\cite{deng2009imagenet} dataset. The model parameters are optimized using stochastic gradient descent with a learning rate of 0.001. We set the imitation parameter $\lambda$ according to the source and target model dependencies as discussed in Section~\ref{sec:lamb}. 

\vspace{0.5em}
\textbf{Baseline Approaches.}
As aforementioned, existing SSDA approaches require access to data from different domains, which are ineffective in federated settings. Besides, none of the existing FL research targets SSDA problems. Although several recent FL research investigated UDA problems~\cite{peng2019federated,peterson2019private,zhuang2021towards}, they target either a single source dataset or a single target dataset for personalization or knowledge aggregation, while our design is under multiple distributed source and target datasets for a more general domain adaptation setting. More importantly, UDA problems assume unknown target ground- truth labels, which cannot effectively extract target domain knowledge from the few labeled target domain samples in SSDA.

Hence, to evaluate the proposed FSSDA, we consider the following two baseline approaches.

\begin{itemize}
\item  \textit{SSDAOnly}: 
Without FL paradigm, individual knowledge cannot be transferred across devices due to privacy concerns. Each device performs SSDA to generate a local target model with its own data but does not participate in federated learning. The accuracy of \textit{SSDAOnly} is derived by averaging the accuracy of target models at end devices.
\item \textit{FLOnly}: Without effective SSDA approaches in federated settings, end devices can only leverage labeled target data to collaboratively learn the target model. There will be no knowledge transfer from the source domain.
\end{itemize}

\begin{table*}[!tbh!] 
\huge
\centering
\resizebox{1\textwidth}{!}
{
\begin{tabular}{c c c c c c c}
\hline
 &{Amazon$\rightarrow$Webcam} & {Amazon$\rightarrow$DSLR} & {Webcam$\rightarrow$Amazon} & {Webcam$\rightarrow$DSLR} & {DSLR$\rightarrow$Amazon} & {DSLR$\rightarrow$Webcam}\\ 
 \cline{1-7}
 \vspace{0.4em}
 SSDAOnly (iid) &66.83\%&66.10\%&56.98\%&75.67\%&49.67\%&73.37\%\\
 \vspace{0.4em}
 FLOnly (iid) &64.08\%&70.10\%&41.07\%&70.10\%&41.07\%&64.08\%\\
 \vspace{0.4em}
  \textbf{FSSDA (iid)} &\textbf{83.01\%}&\textbf{84.94\%}&\textbf{66.23\%}&\textbf{98.45\%}&\textbf{71.39\%}&\textbf{97.63\%}\\
  
  \hline
  \vspace{0.4em}
  SSDAOnly (non-iid) &64.81\%&59.39\%&52.51\%&69.80\%&46.83\%&69.33\%\\
  \vspace{0.4em}
  FLOnly (non-iid) &52.47\%&63.65\%&38.71\%&63.65\%&38.71\%&52.47\%\\
  \vspace{0.4em}
  \textbf{FSSDA (non-iid)} &\textbf{82.15\%}&\textbf{82.15\%}&\textbf{66.09\%}&\textbf{97.20\%}&\textbf{69.67\%}&\textbf{95.48\%}\\

\cline{1-7}
\end{tabular} 
}
\vspace{0.5em}
\caption{Performance comparison among SSDAOnly, FLOnly, and the proposed FSSDA. FSSDA achieves the best results (in bold) for both iid and non-iid cases. }
\label{tab:fssda}
% \vspace{-1em}
\end{table*}
\subsection{Experimental Results}

\textbf{Effectiveness of FSSDA for Single Source.}
We first evaluate the effectiveness of FSSDA for single-source and compare FSSDA with the baseline approaches.
In the experiments, we consider six cases for domain adaptations: Amazon to Webcam, Amazon to DSLR, Webcam to Amazon, Webcam to DSLR, DSLR to Amazon, and DSLR to Webcam, with iid and non-iid data distributions across devices. In all cases, the number of labeled examples is less than the number of unlabeled examples. We conduct each scenario five times and report the average results.

FSSDA leverages both knowledge transfer and knowledge sharing and achieves the best performance for both iid and non-iid cases. 
FLOnly cannot leverage the unlabeled samples, resulting in the worst performance in most cases. Although SSDAOnly leverages unlabeled target domains via knowledge transfer, SSDAOnly cannot utilize the shared knowledge from other end-devices, which makes the learning ineffective. Especially in the non-iid cases, the number of data varies for each end device; the performance degradation of one of the local models affects the aggregated global model.

As shown in Table~\ref{tab:fssda}, \textit{FSSDA outperforms both SSDAOnly and FLOnly in all the cases}. We get the most promising result in the case of Webcam to DSLR both in iid and non-iid settings. The SSDAOnly and FLOnly get around 70\%, whereas our proposed FSSDA methods achieve more than 97\% accuracy.
Both in iid and non-iid settings, DSLR as a target is able to achieve good performance of over 82\% accuracy even when the domain gap is large (Amazon$\rightarrow$DSLR).  
Moreover, due to the domain gap and a large amount of data, it is challenging to transfer knowledge to the Amazon samples (Webcam$\rightarrow$Amazon, and DSLR$\rightarrow$Amazon), but we still achieve better results compared to the baselines. The SSDAOnly and FLOnly can only get around 50\% accuracy, while the FSSDA can achieve accuracy close to 70\%, demonstrating the effectiveness of FSSDA in challenging scenarios.

\begin{figure}[!tb]
\centering
\includegraphics[width=0.9\linewidth]{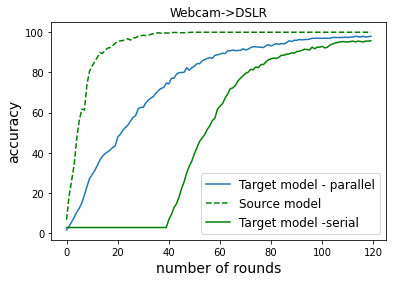}%
% \vspace{-1em}
\caption{Comparison of source, parallel target, and serial target model for Webcam$\rightarrow$DSLR setting.}
\label{fig:para}
% \vspace{-0.5em}
\end{figure}

\vspace{0.5em}

\textbf{Efficiency of FSSDA.}
To evaluate the efficiency of FSSDA, we compare the parallel source and target model (FSSDA) with the serial source and target model (e.g., Webcam$\rightarrow$DSLR, Fig.~\ref{fig:para}). As shown in Fig.~\ref{fig:para} by utilizing parallel training, the proposed FSSDA does not need to wait until the 40th round to adapt the knowledge of the source model. On the other hand, in serial target training, the target model needs to wait until the source model converges \textit{i.e.} first, we train a source model, and after getting a well-trained model, we can start training for the target model. 
Although the source model in the initial stage does not perform well, it still able to promote some knowledge to the target model. Therefore, the parallel training paradigm significantly accelerates the training process and converges faster than the serial training process.

\begin{figure}[!tb]
\centering
\includegraphics[width=0.9\linewidth]{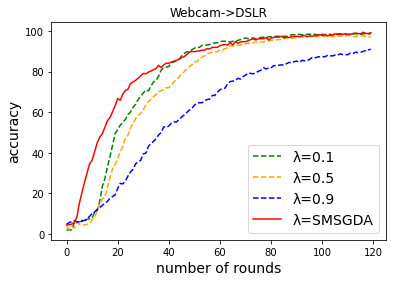}%
% \vspace{-1em}
\caption{Comparison of FSSDA performance for different imitation parameters under a small domain gap (Webcam$\rightarrow$Amazon).}
\label{fig:im}
% \vspace{-1em}
\end{figure}

\begin{figure}[!tb]
\centering
\includegraphics[width=0.9\linewidth]{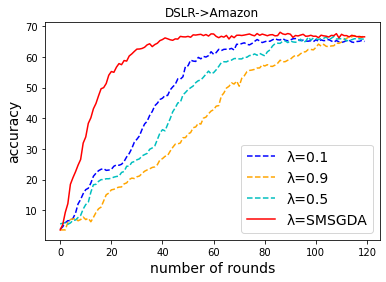}%
% \vspace{-1em}
\caption{Comparison of FSSDA performance for different imitation parameters under a large domain gap (DSLR$\rightarrow$Amazon).}
\label{fig:da}
% \vspace{-1em}
\end{figure}

\vspace{0.5em}
\textbf{Impact of Imitation parameters.}
As we discussed in section~\ref{sec:lamb}, the imitation parameter plays a huge role in transferring knowledge from the source domain to the target domain. We illustrate the impact of the imitation parameter by comparing the performance of FSSDA using different values ($\lambda=0.1, 0.5, 0.9$). A large $\lambda$ indicates learning more knowledge from the target domain (hard label) and less from the source domain (soft label), and vice versa. To evaluate the impact of the imitation parameter, we consider both small domain gap (e.g., Webcam$\rightarrow$DSLR, Fig.~\ref{fig:im}) and large domain gap (e.g., DSLR$\rightarrow$Amazon, Fig.~\ref{fig:da}). We observe that when the domain gap is large (Fig.~\ref{fig:da}), then at the initial stage, a lower value of $\lambda$, \textit{i.e.}, $0.1$ will speed up the performance of the target model, but at the end, performance degrades, which shows the impact of negative transfer. Furthermore, a larger $\lambda$, \textit{i.e.}, $0.9$, finally achieves good accuracy but does not converge quickly compared to the adaptive imitation parameter using SMSGDA. As we see from Fig.~\ref{fig:im} when the domain gap is small, the negative transfer will not be significant ($\lambda=0.1$). Thus we can able to rely more on the source domain. But as we try to learn an adaptive imitation parameter irrespective of the domain difference, in that case, the adaptive imitation parameter using SMSGDA performs well for any situation (small or large domain gap). Overall, compared to all cases ($\lambda=0.1, 0.5, 0.9$), the adaptive imitation parameter using SMSGDA can train the target model faster and converge quickly in both cases.

\vspace{0.5em}
\textbf{Effectiveness of FSSDA for Multi-Source.}
We evaluate the performance of FSSDA under a multi-source scenario. In a single source domain with Amazon as a target, FSSDA can reach close to 70\%, which is lower than the DSLR or Webcam as a target. Amazon has a large amount of data and domain gap compared to the other two domains (DSLR and Webcam). Therefore in the multi-source scenario, we specifically focus on Amazon as a target to improve the overall accuracy while exploring the knowledge from Webcam and DSLR.
We use the same setting as our single source experiment.

\begin{figure}[!tb]
\centering
\includegraphics[width=0.9\linewidth]{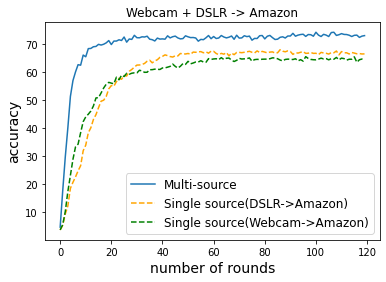}%
% \vspace{-1em}
\caption{Comparison of FSSDA for multi-source scenario (Webcam + DSLR $\rightarrow$Amazon).}
\label{fig:mul}
% \vspace{-1em}
\end{figure}

Our multi-source FSSDA outperforms both single-source results from Webcam and DSLR (considered as the baseline). As shown in Fig.~\ref{fig:mul}, while acquiring the knowledge from multiple sources, the target model not only achieves a better accuracy compared to the single-source model but also speeds up the overall training process to converge faster. The overall average accuracy for Amazon as a target in the multi-source scenario is 74.40\%, which is better than the single source accuracies of 66.23\% (Webcam$\rightarrow$Amazon) and 71.39\% (DSLR$\rightarrow$Amazon).


\section{Conclusion}\label{sec:con}
This paper proposed an innovative approach to achieve SSDA over multiple distributed and confidential datasets, named by Federated Semi-Supervised Domain Adaptation (FSSDA). FSSDA integrates SSDA with federated learning based on the strategically designed knowledge distillation techniques. Specifically, FSSDA includes three key modules: semi-supervised knowledge transfer, parallel training, and adaptive imitation parameter selection, to enable effective and efficient SSDA in a privacy-preserving manner. Moreover, we generalized the proposed FSSDA for multi-source domain scenarios. Finally, we empirically explored SSDA performance under various iid and non-iid federated settings. 
Extensive experimental results on Office dataset validated the effectiveness and efficiency of our design. 

\bibliographystyle{IEEEtran}
\bibliography{reference.bib}

\end{document}